# Ergo:
# A Graphical Environment for Constructing Bayesian Belief Networks

Ingo Beinlich, M.D.
Edward Herskovits, M.D.
*Noetic Systems Incorporated*

## Abstract

We describe an environment that considerably simplifies the process of generating Bayesian belief networks. The system has been implemented on readily available, inexpensive hardware, and provides clarity and high performance. We present an introduction to Bayesian belief networks, discuss algorithms for inference with these networks, and delineate the classes of problems that can be solved with this paradigm. We then describe the hardware and software that constitute the system, and illustrate Ergo's use with several examples.

## Introduction

A graphical representation of probabilistic relationships among variables, called a *Bayesian belief network*, has been independently defined by several researchers. There is thus a plethora of synonyms for Bayesian belief networks [Pearl 1987], such as *causal nets* [Good 1961a; Good 1961b], *probabilistic causal networks* [Cooper 1984], *influence diagrams* [Howard 1984; Shachter 1986], and *causal networks* [Lauritzen 1988]. We shall use the term *belief network* in this paper.

Belief networks provide a conceptual framework for constructing expert systems [Cooper 1989; Horvitz 1988]; they function as platforms for knowledge acquisition [Chavez 1989; Heckerman 1989; Lehmann 1988] and for normative probabilistic inference [Lauritzen 1988; Pearl 1986; Suermondt 1988]. This representational power is particularly helpful when used in complex domains, in which conclusions, intermediate states, and evidence are related by extensive interactions, making knowledge acquisition and knowledge-base maintenance difficult. A belief network clearly specifies the variables, associations, and probabilities relevant to a particular domain, and can thus form the basis for communication and consensus formation among experts [Bonduelle 1987].

Belief networks are primarily characterized as being used for classification; this class of problems excludes planning, yet includes many interesting problems, such as constraint satisfaction, that traditionally have been solved with classical AI methods. Recent research has shown that any influence diagram used for decision making can be transformed to an equivalent belief network [Cooper 1988].

A belief-network is a finite directed acyclic graph in which nodes represent the variables of interest, and arcs from parent nodes to child nodes represent a probabilistic association among the child and its parents. Probabilities are attached to nodes and to arcs in a belief network; these probabilities capture the uncertainty inherent to the relationships among the variables. In particular, for each node $x_i$ with a set of parents $\pi_i$, there is a conditional probability distribution $P(x_i \mid \pi_i)$; for each $x_i$ without parents, there is a prior probability distribution $P(x_i)$.

Conditional probabilities in belief networks can be interpreted as "if-then" rules in the construction of probabilistic expert systems. In this sense, a rule in a belief network is a conditional probability of the form $P(x_i \mid y_1, y_2, \ldots, y_n)$, where $x_i$ and $y_1, y_2, \ldots, y_n$ are variables with known values. Each variable has an associated set of rules, which is the collection of conditional probabilities for each possible combination of values that this variable and its parents can assume.

The prior and conditional probabilities explicitly represented in a belief network are sufficient for computing any probability of the form $P(\Theta \mid \Phi)$, where $\Theta$ and $\Phi$ are members of the power set of this belief network's variables. The key feature of the belief-network paradigm is its explicit characterization of conditional independence among variables, which in turn decreases the number of probabilities required to capture the full joint distribution.

## Inference Algorithms

The practicality of Bayesian reasoning has been debated in the literature for several decades [Buchanan 1985, pp. 235–237; Cooper 1989; Rich 1983]. In particular, slow inference times and the need for a large number of probabilities have been cited. However, several algorithms have been developed during the last ten years that greatly increase the efficiency of Bayesian inference. They share the additional advantage of operating directly on the graphical network structure; thus, the same graph used for knowledge acquisition can be immediately and directly used for validation of that knowledge, leading to rapid model refinement.

Algorithms for belief-network inference, such as those developed by Pearl [Pearl 1986; Suermondt 1988] and Lauritzen and Spiegelhalter [Lauritzen 1988], allow the user to instantiate values for some nodes, after which these algorithms compute posterior distributions over the remaining nodes. These methods thus provide a simple yet general mechanism whereby the user may enter evidence and determine the ensuing implications.

Pearl's algorithm implements a local message-passing system for probability updates in singly connected networks. To cope with multiply connected networks, the method of cutset conditioning has been proposed [Pearl 1988, pp. 204–210] and implemented [Suermondt 1990]. This technique results in a time complexity proportional to the product of the size of the network, the number of cutset instantiations and (without special scheduling techniques) the size of the evidence set. The number of cutset instantiations is exponential in the number of undirected cycles, and for many applications this will result in impractical running times.

The Lauritzen-Spiegelhalter algorithm rearranges the network into a tree by forming clusters of nodes (cliques). The complexity of evidence propagation using this algorithm is linear in the number of cliques, and proportional to the size of the largest clique in the network. The size of a clique is exponential in the number of nodes in that clique. As a node always forms a clique with at least its parents, the maximum number of parents over all nodes in the network is an important determinant of the running time. In contrast to Pearl's algorithm, observing evidence makes inference faster by simplifying the tree of cliques.

Approximate algorithms are also being developed [Chavez 1989; Henrion 1988; Shachter 1989]; they allow the user to trade inference time for precision or accuracy. As the sizes of implemented problems increase, the use of these algorithms will become more prevalent.

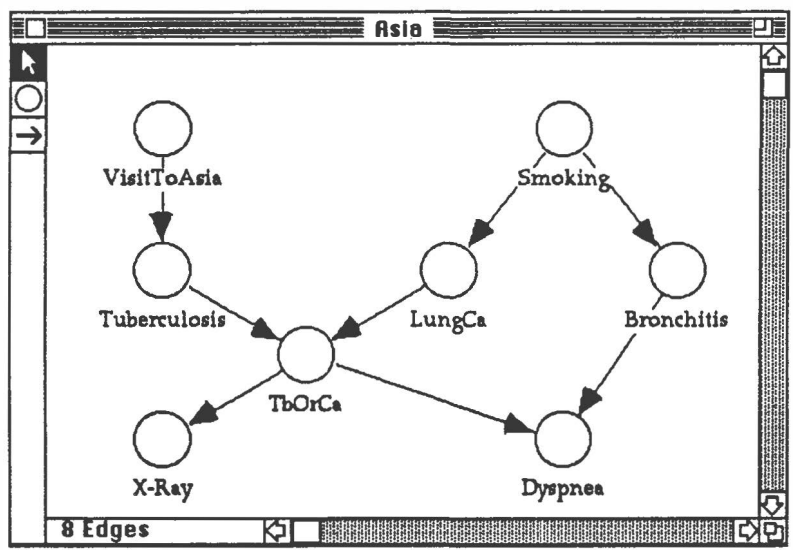

Figure 1

*A graph-editor window in Ergo. The user may edit networks in multiple windows using familiar Macintosh tools*

## Ergo

Although belief networks provide an intuitive medium for knowledge acquisition and inference in probabilistic expert systems, graphical tools are required for the creation and manipulation of any nontrivial network. Ergo provides the necessary environment for rapid prototyping and implementation of probabilistic expert systems. Ergo contains: (1) a graph editor, (2) a probability editor, and (3) a numerical engine.

### *The Graph Editor*

The graph editor provides a drawing environment for manipulating the network structure (Figure 1). Nodes and edges are created and deleted much as they would be in a mouse-based drawing program; nodes (along with their corresponding conditional-probability matrices) and arcs may be copied and pasted either within a win-dow or among windows. This facility allows the user to construct networks incrementally; several subnetworks can be developed individually, and then merged into a larger structure using the copy and paste features. Furthermore, a library of prototypical structures could be designed once and then imported as needed, further modularizing system design and implementation.

*The Probability editor*

At any time during network construction the user may choose to enter probabilities for any node or change its characteristics by using the probability-editor (Figure 2). For each node its name, number of values, and a label for each value is displayed. Probabilities are edited in a tabular format, and can be pasted from other networks or other applications (such as spread-sheets or statistical packages).

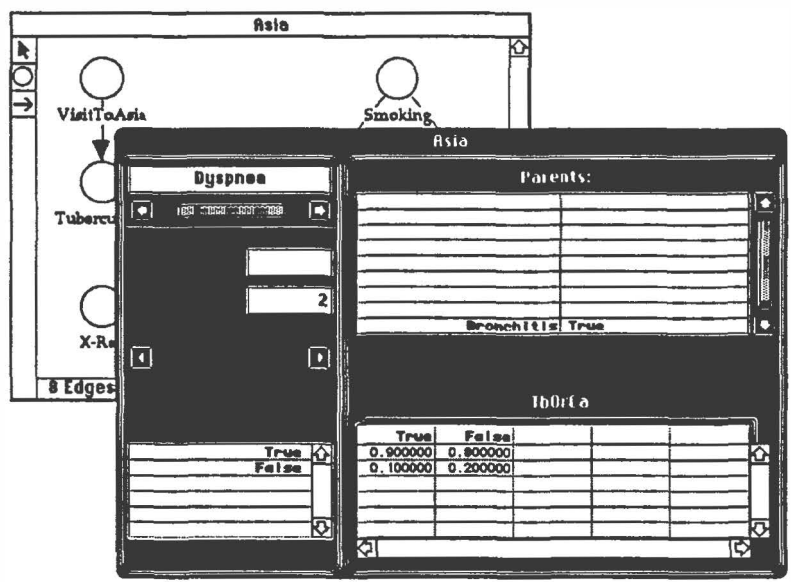

Figure 2

*A probability-editor window in Ergo. For each node its name, number of values, and their labels can be edited. Probabilities are shown with the values of their parents.*

*The Numerical Engine*

A node is instantiated by selecting one of its values from a pop-up menu. This evidence can be propagated automatically, or at the user's request after several instantiations. Results for each node are shown either as histograms or as numerical values in a pop-up menu.

We have implemented an efficient version of the Lauritzen-Spiegelhalter (LS) algorithm. A network is first triangulated using a modified version of maximum cardinality search [Tarjan 1985] that allows for disconnected graphs. The graph is then compiled to a tree (or forest) of cliques as described by Pearl [Pearl 1988, p. 113]. Probability updating in the clique tree follows the scheme described by Lauritzen and Spiegelhalter. Our method of evidence absorption differs from that proposed in their paper in two respects:

- Clique potentials incompatible with new evidence are removed from the cliques and are not set to zero as proposed by Lauritzen and Spiegelhalter. The speed of the following update step increases for any evidence on a variable with more than two values. Consider for example a clique with nodes A, B, and C, where each node has 3 values. This clique has 27 potentials describing its probability distribution. Now assume that node C is observed to have value $c_1$. All potentials in the clique (ABC) with values $c_2$ and $c_3$ for C are incompatible with this evidence and are removed. This step takes at most 27 operations (to check the consistency of each potential). The resulting clique (AB | C=$c_1$) now has only nine potentials. During updating, this clique must propagate its values to its parent and later to its children; the number of steps for propagation is proportional to the number of potentials in the clique. Assume that this clique has only one child (a clique can have at most one parent); propagation then requires 2*9 = 18 steps. The total number of steps involved in updating this clique is 27+18 = 45 steps. In comparison, setting clique potentials to zero involves at most 27 steps, but the clique size remains constant; thus, the update takes a total of 27 + 2* 27 = 81 steps. This benefit increases with the number of children.

- The existing clique tree is modified to accommodate new evidence, instead of being completely rebuilt as Lauritzen and Spiegelhalter suggest. This approach is similar to the join-tree method described by Jensen et al. [Jensen 1988], and obviates an initialization step before every update.

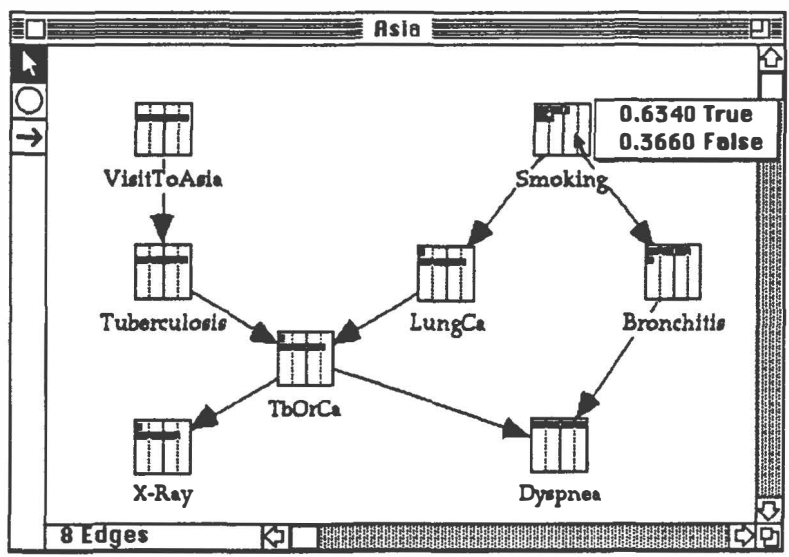

Figure 3

*Inference in Ergo. In this example the node Dyspnea has been instantiated to True. The evidence is propagated through the network and the resulting posterior distributions are shown as histograms. The user may also select any node to view its distribution numerically.*

*Refining the model*

After cycling between network construction and evaluation, the user may save the network to a file, print its structure, or export the graph to any of several Macintosh drawing or writing programs. In addition, the user may choose to export a text-format description of the network so that the file can be read by other programs.

Using a paradigm of iterative construction and evaluation, a group of users can develop subnetworks in parallel and then merge them in a common window. This paradigm owes its feasibility to the modularity conferred on belief networks by the explicit representation of conditional independence.

### Performance

We first describe an implementation of the Asia belief network, which was presented in [Lauritzen 1988], and is shown in our figures. A fictional piece of qualitative knowledge is considered: Patients might have tuberculosis, lung cancer, or bronchitis. These diseases might cause a positive finding on a chest X-ray or shortness of breath (dyspnea). The prior probabilities of the diseases are influenced by a history of smoking and a recent visit to Asia.

Using Ergo on a Macintosh II, we drew and specified the network in a few minutes. Network compilation took less than a second, and resulted in a triangulated graph identical to the full moral graph in the paper by Lauritzen and Spiegelhalter. Evidence absorption and propagation also was instantaneous.

We further tested Ergo with the ALARM network (see Figure 4), which has been used to analyze the complexity and behavior of different inference algorithms [Beinlich 1989]. This network contains 37 nodes (most of them having three or more possible values), and 46 arcs. The process of triangulation and clique produced a clique tree with a total of 1,440 potentials. For a single piece of evidence, the total time to return posterior probability distributions for all 37 nodes required approximately 250 msec. This value is an upper bound on inference time, as the sizes of the cliques in the clique tree *decrease* with increasing size of the evidence set.

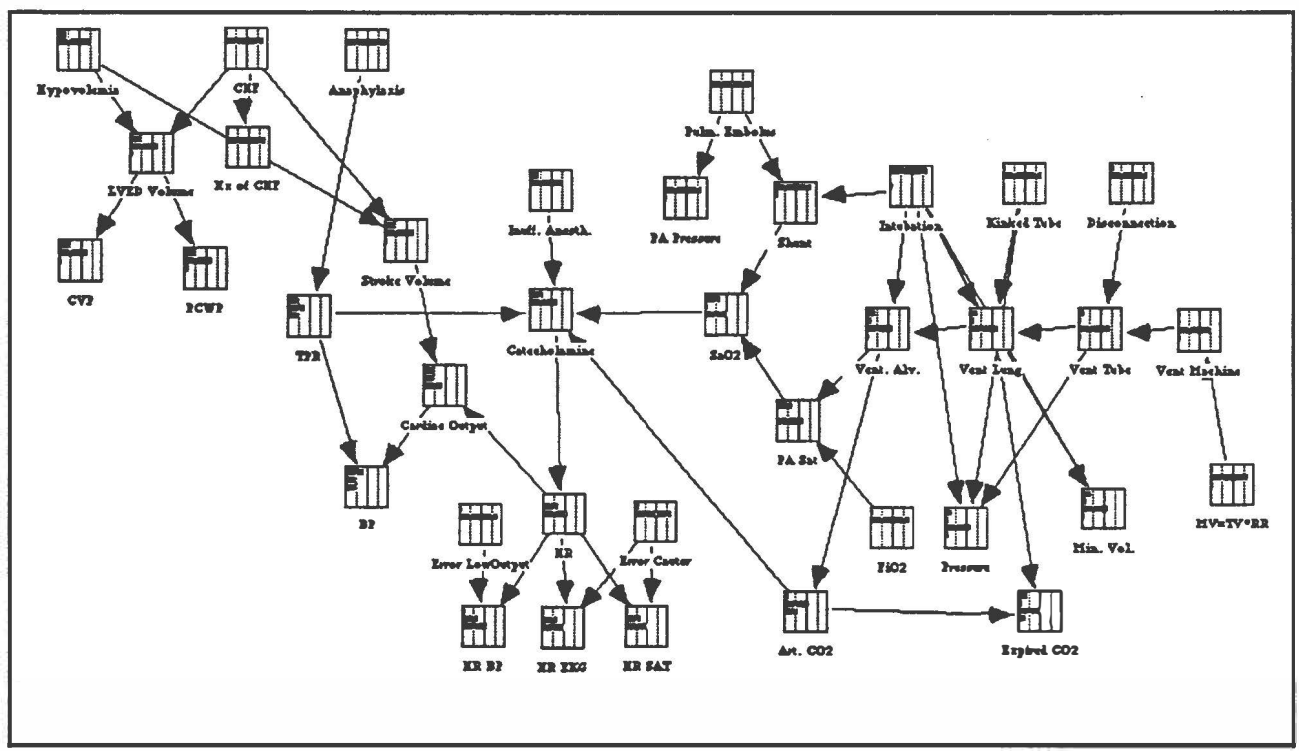

Figure 4 *The ALARM network as created in Ergo*

## Implementation

The numerical engine is written in the C programming language, and is readily portable to any UNIX-based environment as a library that can be linked to other applications. The engine works as a function that takes as input a vector of instantiated nodes, and returns a vector of probability distributions over the remaining nodes. A network produced in Ergo can be saved as a resource for use in other programs.

On the Macintosh such a resource can be incorporated into a Hypercard stack. This resource allows the user to construct, using the Hypertalk programming language, a customized user interface that runs a Bayesian expert system independently of Ergo. The stack fully supports instantiation and inference, and may be distributed to any users with computers that run Hypercard.

## Summary

Ergo demonstrates the feasibility of constructing and using nontrivial Bayesian expert systems; it combines an intuitive user interface with an efficient and portable numeric engine, and is implemented on readily available hardware. These features combine to make Ergo a powerful system for rapid prototyping of belief networks.